# Possibilistic Constraint Satisfaction Problems or "How to handle soft constraints ?"

Thomas Schiex*
CERT–ONERA (DERA–GIA)
2, Av. Edouard Belin
BP 4025
31055 Toulouse Cedex
FRANCE

## Abstract

Many AI synthesis problems such as planning or scheduling may be modelized as constraint satisfaction problems (CSP). A CSP is typically defined as the problem of finding **any** consistent labeling for a fixed set of variables satisfying **all** given constraints between these variables. However, for many real tasks such as job-shop scheduling, time-table scheduling, design..., all these constraints have not the same significance and have not to be necessarily satisfied. A first distinction can be made between hard constraints, which every solution should satisfy and soft constraints, whose satisfaction has not to be certain. In this paper, we formalize the notion of *possibilistic constraint satisfaction problems* that allows the modeling of uncertainly satisfied constraints. We use a possibility distribution over labelings to represent respective possibilities of each labeling. Necessity-valued constraints allow a simple expression of the respective certainty degrees of each constraint.

The main advantage of our approach is its integration in the CSP technical framework. Most classical techniques, such as Backtracking (BT), arc-consistency enforcing (AC) or Forward Checking have been extended to handle possibilistics CSP and are effectively implemented. The utility of our approach is demonstrated on a simple design problem.

## 1 Introduction

There are a lot of publications about constraints, and more specifically in the CSP framework, but most of these papers try to tackle the higly combinatorial nature (NP-Hard) of such problems, only considering *hard* constraints.

This paper gives a clear meaning to what could be a *soft* constraint, how it may be expressed and how soft constraint satisfaction problems may be solved. Our aim is not to give a "general" theoretical framework for expressing soft constraints (such approaches may be found in [Satoh90] using first and second order logic to express preferences or in [Freuder89], relying on a problem space and a general measure on this space), but to give a specific (and hopefully useful) meaning to such constraints leading to "efficient" solving techniques.

Non standard logics are manyfold that allows the expression of probabilities [Nilsson85], or preferences [Shoham87]. In particular, Zadeh's possibility theory [Zadeh78] has already been successfully used for modeling uncertainty and preferences in the frame of propositional and first-order logic by Dubois, Prade and Lang leading to the so-called "possibilistic logic" [Lang91b]. One of the desirable feature of possibilistic semantics is the tolerance to "partial" consistency, which allows a sort of paraconsistent reasoning. Another interesting feature of possibilistic logic is the close relationships between necessity measures and Gärdenfors "epistemic entrechment" relation [Gardenfors et al.88].

The main idea is to encapsulate preferences (or respective certainty degree) among labelings in a "possibility distribution" over labelings. Such a distribution naturally induces two (possibility and necessity) measures over constraints. However, it is not clear how to simply express such a distribution.

A possible answer is to express bounds on necessity (or possibility) measures of constraints, defining a set of possibility distribution among labelings. One can then define a set of "most possible" labelings satisfying these bounds.

The structure of the paper is as follows : the section 2.1 recalls how a Constraint Satisfaction Problem may be defined and which objects are involved ; the section 2.2 presents how a possibility distribution implicitly defines measures on constraints ; the section 2.3 shows how bounds on necessity measures over constraints define "best" labelings.

The next section rapidly presents algorithmic issues for possibilistic CSP solving and shows how specific satisfaction (Backtrack) and consistency enforcing (Arc-consistency [Mackworth77]) techniques may be built, taking into account the induced possibility distribution.

In section 3, we give an example of application of possibilis-tic CSP to a simple design problem. Both representation and solving issues are addressed. Section 4 compares our results with related works and is followed by a presentation of further possible researchs.

---

*e-mail : schiex@cert.fr or schiex@irit.fr

## 2 Possibilistic constraint satisfaction problems

### 2.1 An informal meaning

Let us breafly recall the definition of a classical Constraint Satisfaction Problem as definitions may change among authors.

A typical CSP involves a set $X = \{x_1, \ldots, x_n\}$ of $n$ variables and a set $D$ of $n$ associated domains. Each variable $x_i$ takes its value in its associated domain $d_i$. In the following, we shall restrict ourself to finite domains.

Domains and variables are intrinsically bound together to form what we will call a domain-variable. The domain-variable $i$ is defined as a pair $(x_i, d_i)$ where $x_i$ is a variable and $d_i$ its associated domain. We will call $V$ the set of domain-variables defined by $X$ and $D$.

The fact that some domain-variables take some specific values in their domains will be represented by a labeling. A labeling $l_W$ of a set $W$ of domain-variables is simply defined as an application on $W$ such that :

$$\forall i \in W, l_W(i) \in d_i$$

Alternatively, a labeling $l_W$ will be considered as its map (the set $\{(x, l_W(x))/x \in W\}$).

Further, a set of constraints $C$ is considered. Each constraint $k_i(i_1, \ldots, i_{n_i})$ on the set of domain-variables $V_i = \{i_1, \ldots, i_{n_i}\}$ is a set of labelings of $V_i$.

We will say that a labeling $l_W$ of $W$ *satisfies* a constraint $k_i(i_1, \ldots, i_{n_i})$ (noted $l_W \models k_i$) iff $V_i \subset W$ and $\exists l \in k_i(i_1, \ldots, i_{n_i})/l \subset l_W$.

We will say that a labeling is *complete* iff it is defined on $V$, it will be *partial* otherwise.

**Definition 2.1** *Let us consider $l_A$ and $l_B$ two partial labelings ($A \subset V$, $B \subset V$). We will say that $l_A$ is more defined than $l_B$ (noted $l_A \succeq l_B$) iff $l_B \subset l_A$.*

A partial labeling $l$ typically represents the set of every complete labeling that are more defined than $l$.

We finally define the following algebra over constraints :

- For any given constraint $k_i(i_1, \ldots, i_{n_i})$, we will note $\neg k_i(i_1, \ldots, i_{n_i})$ the constraint on $V_i$ that is unsatisfied when $k_i$ is satisfied. $\neg k_i(i_1, \ldots, i_{n_i})$ is simply the complement of $k_i(i_1, \ldots, i_{n_i})$ in the set $L_{V_i}$ of every labelings over $V_i$ ;

- Given two constraints $k_i$ and $k_j$, we will note $k_i \wedge k_j$ the constraint on $V_i \cup V_j$ that is satisfied when both $k_i$ and $k_j$ are satisfied. $k_i \wedge k_j$ is simply defined as the set of every labeling $l$ over $V_i \cup V_j$ such that $l \models k_i$ and $l \models k_j$ ;

- Given two constraints $k_i$ and $k_j$, we will note $k_i \vee k_j$ the constraint on $V_i \cup V_j$ that is satisfied when one of $k_i$ or $k_j$ are satisfied. $k_i \vee k_j$ is simply defined as the set of every labeling $l$ over $V_i \cup V_j$ such that $l \models k_i$ or $l \models k_j$ ;

The usual problem is then to find a labeling of the domain-variables in $V$ that satisfies the conjunction of every constraint in $C$. Cryptograms (such as SEND+MORE=MONEY, LYNDON*B=JOHNSON), the $n$ queens problem,... are typical instances of CSP.

### 2.2 Modeling soft constraint with possibility and necessity measures

In classical first order logic, soft constraint may be formalized through interpretation ordering [Satoh90]. Possibilistic CSP, as indicated by their name, relies on a possibility distribution over labelings.

Let $L_V$ be the finite set of all possible labeling of the domain-variables in $V$. A *possibility distribution* on $L_V$ is a function $\pi$ from $L_V$ to $[0, 1]$. $\pi$ is said to be *normalized* if and only if $\exists l \in L_V$ such that $\pi(l) = 1$. We define the *sub-normalization degree* of $\pi$ as the quantity $SN(\pi) = 1 - \text{Sup}(\{\pi(l)/l \in L\})$. Obviously, $SN(\pi) = 0$ iff $\pi$ is normalized.

Let $K$ be the set of every possible constraints on any non-empty subset of $V$. The possibility distribution $\pi$ on $L_V$ induces two functions on $K$ called possibility and necessity measures respectively noted $\Pi_\pi$ and $N_\pi$ defined as follows :

$$\Pi_\pi : K \to [0, 1]$$
$$\forall k \in K,$$
$$\Pi_\pi(k) = \text{Sup}(\{\pi(l), l \in L_V, l \models k\} \cup \{0\})$$

$$N_\pi : K \to [0, 1]$$
$$\forall k \in K,$$
$$N_\pi(k) = \text{Inf}(\{1 - \pi(l), l \in L_V, l \models \neg k\} \cup \{1\})$$
$$N_\pi(k) = 1 - \Pi_\pi(\neg k)$$

Let us denote by $\bot$ any unsatisfiable constraint (there is no $l \in L_V/l \models \bot$ i.e, $\bot$ is a constraint that contains no labeling) and by $\top$ the ever satisfied constraint (i.e, the set of all labelings on $V$, $L_V$).

- $N_\pi(\top)$ is obviously equal to 1 i.e., ever satisfied constraints are satisfied ;

- $N_\pi(\bot) = SN(\pi)$ which is generally not equal to 0 ! This means that unsatisfiable constraint may be somewhat required to be satisfied. This is dependant upon the fact that the possibility distribution $\pi$ is not required to be normalized. This choice has been made to cope with partial inconsistencies.

- $\forall k_1, k_2 \in K,$
  $N_\pi(k_1 \wedge k_2) = \text{Inf}(\{1 - \pi(l), l \in L_V, l \not\models k_1 \wedge k_2\} \cup \{1\})$

$$= \text{Inf}(\{1 - \pi(l), l \in L_V, l \not\models k_1\}$$
$$\cup \{1 - \pi(l), l \in L_V, l \not\models k_2\} \cup \{1\})$$
$$= \text{Inf}(\{N_\pi(k_1), N_\pi(k_2)\}).$$

The possibility $\Pi_\pi(k)$ represents what its name suggests i.e., the possibility for the constraint $k$ to be satisfied according to the knowledge of reference. The necessity $N_\pi(k)$ tends towards 1 when the possibility that $k$ being unsatisfied tends toward 0, measuring to what extent the satisfaction of $k$ is entailed by the knowledge of reference (given by $\pi$).

Clearly, possibilistic CSP, as possibilistic logic, is not meant to express fuzzy constraints[1] as measures are attached to precise constraints. The statement "It is 0.7 necessary that the product be delivered before the 21th" may be translated in possibilistic CSP to something like $N_\pi(D \leq 21) = 0.7$ where $D$ is the variable corresponding to the delivering day. Possibilistic CSP is not intended to modelize a statement such as "The product should be delivered not too late after the 21th"[2].

Because of the *min* and *max* operators, the precise values of necessity or possibility is not so important. The essential is the total pre-order induced by them. Thus, necessity degrees express **preference degrees**, $N_\pi(k) > N_\pi(k')$ expressing that the satisfaction of $k$ is preferred to the satisfaction of $k'$. Therefore, possibilistic CSP are closely related to Hierarchical CSP as described in the frame of Constraint Logic Programming in [Borning et al.89].

### 2.3 Possibilistic CSP : definition and semantics

The only difference between a classical and a possibilistic CSP is the introduction of necessity-valued constraint instead of simple constraint. A necessity valued constraint is a pair :

$$(k_i(i_1, \ldots, i_{n_i}), \alpha)$$

where $k_i(i_1, \ldots, i_{n_i})$ is a classical constraint and $\alpha \in [0, 1]$.

A typical possibilistic CSP is then defined by a finite set $X$ of variables, a finite set $D$ of associated finite domains (defining a set $V$ of domain-variables) and by a finite set $C$ of necessity valued constraints. It will be noted either $(X, D, C)$ or $(V, C)$.

The necessity-valued constraint $(k, \alpha)$ expresses that $N_\pi(k) \geq \alpha$ i.e., that the satisfaction of $k$ is at least $\alpha$-necessary. The necessity-valued constraint $(k, 1)$ expresses that $k$ should absolutely be satisfied, and therefore takes the usual meaning of a constraint in classical CSP ; $(k, 0)$ is totally useless as it expresses that the necessity measure of $k$ should be at least 0, which is always true.

---

[1] Vague relations seen as a fuzzy set of authaurized labeling. See [MC92,Dubois et al.89,Rosenfeld et al.76].

[2] In fact, such a predicate may be decomposed in a set of crisp predicates ($\alpha$-cuts of the vague constraint). In our case, the domains being finite, the set of $\alpha$-cuts is finite and a given fuzzy constraint may be decomposed in a finite set of possibilistic constraints. However, this (possibly automatic) conversion may be heavy and the result is (from an expressive view-point) quite distant from the original knowledge.

The notion of "constraint satisfaction" will now depend on a possibility distribution $\pi$ on $L_V$. Let us consider $(k, \alpha)$ a necessity-valued constraint.

We will say that $(k, \alpha)$ is satisfied by $\pi$ (noted $\pi \models (k, \alpha)$) iff the necessity measure $N_\pi$ induced by $\pi$ on $K$ verifies $N_\pi(k) \geq \alpha$. Considering the whole constraint set $C$, we will say that the CSP $(V, C)$ is satisfied by a possibility measure $\pi$ iff the necessity measure induced by $\pi$ verifies :

$$\forall (k, \alpha) \in C, N_\pi(k) \geq \alpha$$

Thus a possibilistic CSP has not a set of consistent labeling, but a set of possibility distributions on the set of all labelings on $V$.

- If we consider a specific distribution $\pi$, the most possible labeling will have a possibility equal to $(1 - SN(\pi))$ ;
- On another hand, if we consider a specific (complete) labeling $l$, its compatibility with the knowledge of reference (noted $\mathfrak{C}(l)$) will be the maximum of $\pi(l)$ for every $\pi$ which satisfies $C$. Its incompatibility (noted $\mathfrak{I}(l)$) will be its complement to 1.

Thus the degree of consistency of the possibilistic CSP or $\mathbb{C}(V, C)$ may be defined as the maximum of $1 - SN(\pi)$ for every $\pi$ which satisfies $C$ or, equivalently, as the compatibility of the most compatible labeling. Its inconsistency degree $\mathbb{I}(V,C)$ will be the complement to 1 of its consistency degree :

- $\mathbb{C}(V, C) = \text{Sup}_{\pi \models C}(\text{Sup}_{l \in L_V}(\pi(l)))$
  $= \text{Sup}_{\pi \models C}(1 - SN(\pi))$
  $= \text{Sup}_{l \in L_V}(\mathfrak{C}(l))$ ;
- $\mathbb{I}(V, C) = 1 - \mathbb{C}(V, C)$
  $= \text{Inf}_{\pi \models C}(SN(\pi))$
  $= \text{Inf}_{\pi \models C}(N_\pi(\bot))$
  $= \text{Inf}_{l \in L_V}(\mathfrak{I}(l))$ ;

Thus, the inconsistency degree of a possibilistic CSP is equal to the smallest necessity degree of the unsatisfiable constraint $\bot$ for all possibility distribution satisfying $C$.

The computation of the inconsistency degree of a possibilistic CSP is made easier by the fact that one can define a maximal possibility distribution among all possibility distribution satisfying $(V, C)$.

**Theorem 2.1** *Let $\mathcal{P} = (V, C)$ be a possibilistic CSP, we define the possibility distribution $\pi_\mathcal{P}^*$ on $L_V$ by :*

$$\forall l \in L_V, \pi_\mathcal{P}^*(l) = \text{Inf}_{(k_i, \alpha_i) \in C}(\{(1 - \alpha_i)/l \models \neg k_i\} \cup \{1\})$$

*Then for any possibility distribution $\pi$ on $L_V$, $\pi$ satisfies $\mathcal{P}$ iff $\pi \leq \pi_\mathcal{P}^*$.*

Proof :

$\pi$ satisfies $\mathcal{P}$

iff $\forall (k_i, \alpha_i) \in C$, $\pi$ satisfies $(k_i, \alpha_i)$
iff $\forall (k_i, \alpha_i) \in C, N_\pi(k_i) \geq \alpha_i$

iff $\forall (k_i, \alpha_i) \in C, \text{Inf}(\{1 - \pi(l)/l \models \neg k_i\} \cup \{1\}) \geq \alpha_i$
iff $\forall (k_i, \alpha_i) \in C, \forall l \models \neg k_i, \pi(l) \leq 1 - \alpha_i$
iff $\forall l \in L_V, \pi(l) \leq \text{Inf}_{(k_i, \alpha_i) \in C}(\{1 - \alpha_i/l \models \neg k_i\} \cup \{1\})$
iff $\forall l \in L_V, \pi(l) \leq \pi_\mathcal{P}^*(l).\square$

**Corollary 2.1** *We simply conclude that :*

- $\mathfrak{C}(l) = \pi_\mathcal{P}^*(l)$ ;

- $\mathfrak{I}(l) = (1 - \pi_\mathcal{P}^*(l))$ ;

- $\mathbb{C}(\mathcal{P}) = Sup_{l \in L_V}(\pi_\mathcal{P}^*(l))$
  $= 1 - SN(\pi_\mathcal{P}^*)$
  $= Sup_{l \in L_V}(\{Inf_{(k_i, \alpha_i) \in C}(\{(1 - \alpha_i)/l \models \neg k_i\}$
  $\cup \{1\})\})$

- $\mathbb{I}(\mathcal{P}) = 1 - Sup_{l \in L_V}(\pi_\mathcal{P}^*(l))$
  $= SN(\pi_\mathcal{P}^*)$
  $= Inf_{l \in L_V}(Sup_{(k_i, \alpha_i) \in C}(\{\alpha_i/l \models \neg k_i\} \cup \{0\})\})$

Proof : The two first points are immediate.

According to theorem 2.1, we know that :

$\forall \pi$ that satisfies $\mathcal{P}, \pi \leq \pi_\mathcal{P}^*$, i.e.,
$\forall \pi$ that satisfies $\mathcal{P}, \forall l \in L_V, (1 - \pi(l)) \geq (1 - \pi_\mathcal{P}^*(l))$
$\forall \pi$ that satisfies $\mathcal{P}, SN(\pi) \geq SN(\pi_\mathcal{P}^*)$.

So :

$\mathbb{I}(\mathcal{P}) = \text{Inf}_{\pi \models C}(SN(\pi))$
$= SN(\pi_\mathcal{P}^*)$
$= \text{Inf}_{l \in L_V}(Sup_{(k_i, \alpha_i) \in C}(\{\alpha_i/l \models \neg k_i\} \cup \{0\})\})$.

The corresponding result for the consistency degree is immediate.$\square$

Then, computing the inconsistency degree of a CSP means computing the sub-normalization degree of the distribution $\pi_\mathcal{P}^*$. The set of all labeling $L_V$ being finite, we can define the set $L_V^*$ of all labelings of $V$ such that $\forall l^* \in L_V^*, \pi_\mathcal{P}^*(l) = 1 - SN(\pi_\mathcal{P}^*)$. This will be called the set of the best labelings of $\mathcal{P}$. Its elements are the most compatible labelings with the CSP $\mathcal{P} = (V, C)$ among every labeling. The problem of finding a "best labeling" may be reduced to find a labeling $l^*$ that solve any of the following equivalent Min-Max optimisation problems :

$\mathbb{C}(V,C) = Sup_{l \in L_V}(\text{Inf}_{(k_i, \alpha_i) \in C}(\{1 - \alpha_i/l \models \neg k_i\} \cup \{1\}))$
$\mathbb{I}(V,C) = \text{Inf}_{l \in L_V}(Sup_{(k_i, \alpha_i) \in C}(\{\alpha_i/l \models \neg k_i\} \cup \{0\}))$

Such problems may be tackled through many classical tree-search algorithms, namely Depth first Branch and Bound (DFBB), $\alpha - \beta$, or SSS*...

### 2.4 Extending classical CSP algorithms

#### 2.4.1 Generate and Test

The more obvious algorithm to solve classical CSP is the "generate and test" algorithm. It traverses the domain-variables in a predetermined order $(1, \ldots, n)$. In the tree explored, each node corresponds to a labeling. The root of the tree is the empty labeling, the sons of a node $l$ are obtained by extending the labeling $l$ on $(1, \ldots, i)$ with a new variable $i + 1$ and every possible label in $d_{i+1}$. The leafs of the trees are complete labelings that may (or not) satisfy every constraint. In a depth first exploration of the tree, the first labeling that satisfies every constraint is retained.

The corresponding approach in possibilistic CSP will be an optimization problem on the same tree. For each leaf of the tree, we may compute the value of $\pi_\mathcal{P}^*$ on the corresponding complete labeling. In a depth first exploration of the tree, we will retain the set of the labelings that maximize $\pi_\mathcal{P}^*(l)$.

#### 2.4.2 Test and Generate

The next step towards sophistication (and efficiency) is the "test and generate" approach, often referred as the "Back-track" algorithm (BT). The obvious idea is to cut each branch that will necessarily lead to complete labelings that do not satisfy every constraint. Each non-terminal node corresponds to a partial labeling $l$. To possibly lead to a complete labeling that satisfies every constraint, a partial labeling should be *consistent* :

**Definition 2.2** *Given a classical CSP $(V, C)$, a partial labeling $l_A$ on the set of domain-variables $A \subset V$ will be consistent iff $\forall k_i \in C$ such that $V_i \subset A, l_A \models k_i$.*

If a partial labeling $l$ looses its consistency property, every labeling $l'$ more defined than $l$ will also be non-consistent. In the case of complete labeling, non-consistency is equivalent with non-satisfaction.

In the case of a depth first tree exploration, the property of consistency is simply checked at each node. Backtrack occurs when it is not verified. One should note that if a labeling $l$ on $\{1, \ldots, i\}$ is consistent, each labelings $l'$ on $\{1, \ldots, i, i + 1\}$ is consistent iff it satisfies the constraints $k_j$ such that $V_j \subset \{1, \ldots, i, i + 1\}$ and $V_j \not\subset \{1, \ldots, i\}$.

In the framework of possibilistic CSP, we extend the notion of compatibility to partial labelings.

**Definition 2.3** *The compatibility of a partial labeling $l_A$ on $A$ is defined as the maximum of the compatibility of every complete labeling more defined than $l_A$ :*

$$\mathfrak{C}(l_A) = Sup(\{\mathfrak{C}(l)/l \in L_V, l \succeq l_A\})$$

Our aim will be to compute, for each node (i.e. each partial labeling) an upper bound on the compatibility of the partial labeling. An easily computed upper bound of this value is given by :

$$\text{Inf}_{(k_i, \alpha_i) \in C}(\{(1 - \alpha_i)/V_i \subset A, l_A \models \neg k_i\} \cup \{1\})$$

This bound is *exact* for complete labelings. Moreover, it may be incrementally computed as the tree is traversed downwards : if a labeling $l$ on $A$ has been granted an upper-bound $\beta$, each labeling $l'$ on $A \cup \{(x_j, d_j)\}$, more defined than $l$ is granted the upper-bound $\beta'$ :

$$\beta' = \text{Inf}(\{\beta\} \cup \{(1-\alpha_i) \ / \ (k_i, \alpha_i) \in C, \\ V_i \subset A \cup \{(x_j, d_j)\}, \\ V_i \not\subset A, \\ l' \models \neg k_i\})$$

For the sake of clarity, given an explicit ordering $O = (1, \ldots, n)$ on the domain-variables, we will note $C_j^{j+1}$ the set of the necessity valued constraints $(k_i, \alpha_i)$ such that $V_i \subset \{1, \ldots, j, j+1\}$ and $V_i \not\subset \{1, \ldots, j\}$. If we note $\beta$ the upper-bound previously computed on a labeling $l$ on $\{1, \ldots, j\}$ and $l'$ a labeling on $\{1, \ldots, j+1\}$, more defined than $l$, we may compute the upper-bound $\beta'$ on the compatibility of $l'$ via :

$$\beta' = \text{Inf}(\{\beta\} \cup \{(1-\alpha_i)/(k_i, \alpha_i) \in C_j^{j+1}, l' \models \neg k_i\})$$

This decreasing bound is used in a DFBB algorithm to compute one (or every) best labeling. The algorithm simply starts with the empty labeling and extends it according to the vertical ordering $O$. It maintains two parameters of importance : the number $\alpha$ under which a cutoff should take place (increased each time a complete labeling with an augmented compatibility has been found. It should be initially set to zero to ensure optimality, a cutoff takes place as soon as $\beta \leq \alpha$) and the number $\beta$ over which no improvement is possible (the bound on the compatibility of the current partial labeling. It should initially be set to 1). These two bounds offer a great deal of flexibility :

- Typically, if a labeling whose possibility is lower than $a$ is considered as useless, the algorithm should be called with $\alpha$ set to $a$, allowing a more efficient pruning of the tree ;
- On the opposite side, if a labeling of possibility $b$ is considered as enough, the algorithm should be called with $\beta$ set to $b$, allowing the algorithm to stop as soon as a $\beta$ consistent labeling has been found.

Naturally, in the first case we may fail to find a best labeling if its consistency degree is lower than $\alpha$ ; in the second case, we have no garantee that the best labeling has been found. Alternatively, one may stop the algorithm execution upon any event (time exhausted,...) and get the best labeling found up to the occurrence of the event (getting closer to an "anytime algorithm" [Dean et al.88]).

Every usual vertical heuristic ordering (max. cardinality, max. degree,...), may be applied to this tree search. An horizontal heuristic is given by the current bounds obtained for the various labels, but its efficiency is yet to be evaluated (it is well known that horizontal ordering has a strong influence on the efficiency of Min-Max problems solving, e.g. in Alpha-Beta algorithm applied to games [Pearl85]).

### 2.4.3 Consistency enforcing

A further step is to extend the various local consistency notions (node, arc, path and $k$ consistency [Mackworth77,Montanari74]) and their corresponding filtering algorithms[Mohr et al.86,Deville et al.91].

A classical n-ary CSP is said to be arc-consistent iff for every domain-variable $j$, the domain $d_j$ is not empty, and for every label $\nu \in d_j$, for every constraint $k_i$ such that $j \in V_i$, there is a labeling $l$ on $V_i$, more defined than the labeling $\{(j, \nu)\}$, that satisfies $k_i$.

The algorithm that converts a CSP in an equivalent[3] arc-consistent CSP (if it exists) is usually embodied in a single procedure *Revise*, applying to a domain-variable $j$ and a constraint $k_i$ $(j \in V_i)$, that suppresses every label in the domain $d_j$ that does not satisfy the previous property. This procedure is applied repeatedly on the whole CSP until stabilization (AC1 to AC3).

In our case, such a label may still be possible if the constraint $k_i$ is not 1-necessary. In general, the knowledge we may extract is an upper bound on the compatibility of the partial labelings that maps a single variable to a label.

If we consider a variable $j$ and a (non-unary) constraint $k_i$ such that $j \in V_i$ and if we note $U_{V_i}$ the set of unary constraint on any of the variables in $V_i$, the upper bound[4] on the compatibility of the partial labeling $\{(j, \nu)\}, \nu \in d_j$, taking into account $k_i$ and every unary constraint in $U_{V_i}$ is equal to :

$$b(\{(j,\nu)\}, k_i) = \text{Sup}_{l' \in L_{V_i}, l' \supseteq \{(j,\nu)\}} \\ (\text{Inf}_{(k_n, \alpha_n) \in (U_{V_i} \cup \{(k_i, \alpha_i)\})} \\ (\{(1-\alpha_n)/l' \models \neg k_n\} \cup \{1\}))$$

A possibilistic CSP will be said arc-consistent if for every domain-variable $j$, the domain $d_j$ is not empty, and for every label $\nu \in d_j$, the compatibility of $\{(j, \nu)\}$ with respect to the possibilistic CSP $(\{j\}, U_{\{j\}})$ is strictly positive and equal to the minimum of the $b(\{(j, \nu)\}, k_i)$ for every $k_i$ such that $j \in V_i$.

More precisely, if we note $\mathcal{P}_j$ the CSP defined by $(\{j\}, U_{\{j\}})$, a possibilistic CSP is $\delta$-arc-consistent if it is arc-consistent, and :

$$\delta = \text{Inf}_{j \in V}(\text{Sup}_{\nu \in d_j}(\pi^*_{\mathcal{P}_j}(\{(j,\nu)\})))$$

It may be shown that $\delta$ is an upper-bound on the overall consistency $\mathbb{C}(V, C)$.

The main idea to convert a possibilistic CSP into an equivalent[5] arc-consistent possibilistic CSP is then to add unary necessity-valued constraints (rather than suppressing labels) reflecting this bound and to take these new unary constraints in account when the process is repetedly applied.

The *Revise*$_\pi$ procedure we have defined not only filters out necessarily inconsistent labels, but also compute for each label $\nu \in d_j$ the upper bound $b(\{(j, \nu)\}, k_i)$ on the compatibility of the partial labeling $l$ that maps the variable $j$ being filtered to this label $\nu$ taking into account the constraint $k_i$. This bound $b$ may be simply encoded in the CSP by adding a simple unary constraint[6] on $j$ indicating that this label is forbidden with a necessity $1 - b(l, k_i)$.

---

[3] Two CSP $\mathcal{P}_1$ and $\mathcal{P}_2$ are equivalent if they have the same set of solutions, i.e, $\forall l, l \models \mathcal{P}_1 \Leftrightarrow l \models \mathcal{P}_2$.

[4] It is precisely the compatibility of the labeling $\{(j, \nu)\}$ in the CSP $(\{j\} \cup V_i, \{k_i\} \cup U_{V_i})$.

[5] Two possibilistic CSP $\mathcal{P}_1$ and $\mathcal{P}_2$ are equivalent if they have the same set of satisfying possibility distributions, i.e. $\forall \pi, \pi \models \mathcal{P}_1 \Leftrightarrow \pi \models \mathcal{P}_2$, or equivalently $\pi^*_{\mathcal{P}_1} = \pi^*_{\mathcal{P}_2}$.

The additionnal information obtained is taken into account in the tree search algorithm and may greatly enhance the performances of the algorithm (tighter bounds on partial inconsistencies are obtained earlier). The termination (which is quite trivial) and complexity of the algorithm, the unicity of the problem obtained are yet to be formaly determined.

Limited applications of the $Revise_\pi$ procedure during the tree search exploration (so-called Forward Checking, or Partial Look-Ahead) that are usual in CSP are immediately usable in possibilistic CSP and have been implemented.

It should be noted that a possibilistic CSP containing only HARD constraints is strictly equivalent to a classical CSP and that extended algorithms (tree exploration, arc-consistency) behave *exactly* as corresponding classical CSP algorithms. Therefore "softness" costs (almost) nothing when left unused. The only overhead is due to the manipulation of the floating point numbers 1.0 and 0.0 and the operators min/max instead of boolean true and false and logical operators and/or.

## 3 A design problem

A great restaurant want to offer to its clients a computer aided menu designer. The system should integrate "knowhow" knowledge and customer desires to compose a "best menu"composed of a drink (white or red wine, beer or water), an entrance (smoked salmon, caviar, "foie gras", oysters or nothing), a main dish (fish of the day, leg of wild boar, sauerkraut) and a dessert (apple-pie, strawberry ice, fruit or nothing).

We shall first consider the following knowledge :

- The sauerkraut should be accompanied by a beer (a, 0.8), white whine may be possibly considered (b, 0.3), or even water (c, 0.2) ;
- Fish may not be eaten twice in the menu (caviar and oysters will be considered as "fishes") (d, 0.7), and should be accompanied with white wine (e, 0.9) or water (f, 0.2) ;
- Meat should (almost) certainly be eaten with a red wine (g, 0.9) ;
- Foie gras should be accompanied by a soft white wine (h, 0.9) ;
- After the leg of wild boar, a strawberry ice as very good digestive effects (i, 0.5) ;
- No entrance or no dessert is not appreciated (by the restaurant) (j, 0.4), having both no entrance and dessert is even less appreciated (k, 0.6) ;
- Having water as a drink is no good (l, 0.5) ;

---
[6]One may also define $\gamma$-weak arc-consistency enforcing by limiting the $Revise_\pi$ to the inference of unary constraints whose necessity is greater or equal than $\gamma$. 1-weak arc-consistency leads to label suppression. 0-weak arc-consistency is possibilistic arc-consistency.

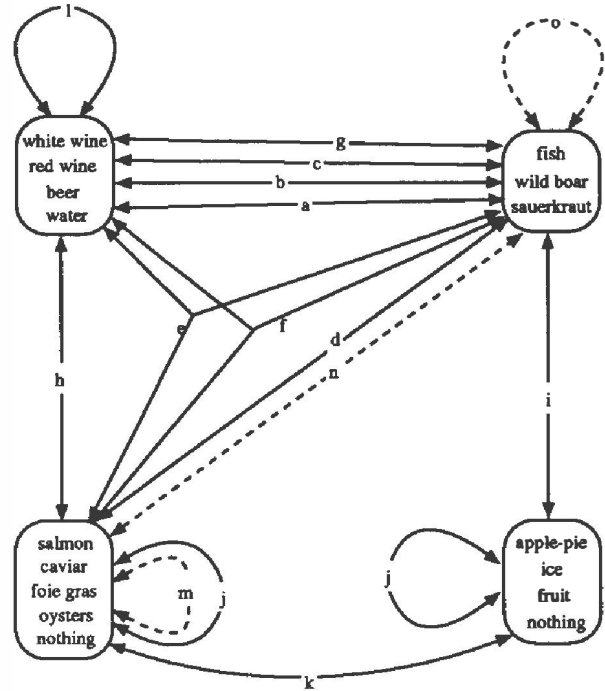

Figure 1: Gastronomic CSP hypergraph

Our client now integrate its preferences :

- I surely do not want any oysters in my menu (m, 1.0) ;
- I would like to eat some fish (n, 0.8) ;
- I would like to taste the sauerkraut (o, 0.2) ;

The encoding in a 4 variables possibilistic CSP is immediate (Cf. figure 1). The basic constraints are represented by the continuous arcs, the client constraints are represented by dotted arcs). The tree explored with the previously outlined "DFBB" algorithm using the ordering *(Dish, Drink, Entrance, Dessert)* is given figure 2. Labels are given by their capitals, cutoffs are indicated by thick lines. The first "best menu" found (compatibility 0.8) is as follows :

- main dish : Fish of the day ;
- drink : White wine ;
- entrance : Foie gras ;
- dessert : Apple-pie ;

The overall consistency degree of the CSP is therefore equal to 0.8. As the knowledge introduced makes no difference between a soft and a dry white wine, our customer will either drink its "foie gras" with a dry wine or its fish with a soft wine. Nobody is perfect...

The size of this problem makes arc-consistency enforcing and forward-checking useless. Nevertheless, one may note that the problem is actually 0.8 arc-consistent. As an example, one of the unary constraint infered by arc-consistency enforcing is a constraint that forbids the label "white vine" with a 0.2 necessity (as $b(\{(\text{drink}, \text{white wine})\}, a) = 0.8$).

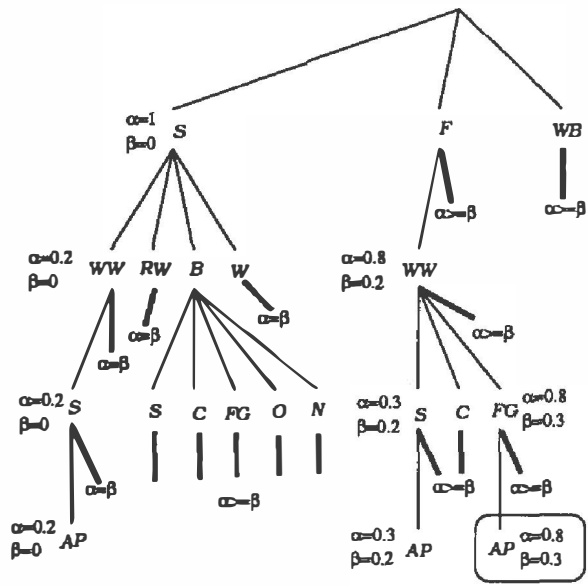

Figure 2: DFBB search

We are currently trying to apply these techniques in the frame of job-shop scheduling. It is clear that the particular nature of the constraints that appears in this framework could (and should) be taken into account in the propagation process, as it may be done in the AC-5 [Deville et al.91] algorithm in classical CSP.

## 4 Related works

The obviously related work is "possibilistic logic" [Lang91b] which has been a fundamental basis for possibilistic CSP definition. J. Lang [Lang91a] has applied propositional possibilistic logic to constraint satisfaction problems. In our opinion, our approach offers greater expressive power (let us recall that the encoding of the SEND+MORE=MONEY problem in propositional logic leads to no more than 2060 clauses and 88 propositional variables) and more varied and powerful techniques (the only resolution technique used in propositional possibilistic logic being essentialy a "backjumping-like" algorithm [Oxusoff et al.89,Gashnig79]).

Other related works include Hierarchical Constraint Logic Programming [Borning et al.89] that allows the expression of prioritized constraints in the body of an Horn clause. Satoh [Satoh90] proposes a formalisation of soft constraints based on an interpretations ordering but does not provide any algorithmic issue.

The system GARI [Descottes et al.85] which is more oriented towards production rules is very close to ours as it compute a solution that is the best compromise under a set of antagonist constraints. It is also close to the OPAL scheduling system [Bel et al.89,Bel et al.88] which has been extended to take in account fuzzy antagonist temporal constraints.

## 5 Further researchs

We are currently working on the conversion of the possibilistic AC1 like algorithm to more sophisticated schemes as AC4 [Mohr et al.86]. A matter of study is also the fix-point semantics of the possibilistic arc-consistency as is has been done for classical CSP [Gusgen et al.88].

Several extension of possibilistic CSPs may be considered :

- Many CSP techniques (AC-n, path or k-consistency, backjumping, learning, tree clustering, cycle cutset) and useful properties (Freuder theorems [Freuder82]) should be adapted or extended to possibilistic CSP ;

- The integration of fuzzy constraints (defined as a fuzzy set of authorized labelings) is almost immediate and leads to an even greater expressive power.

- As is has been shown for possibilistic logic [Dubois et al.91b], the pre-order induced by necessity-valued constraints is a numerical "epistemic entrenchment" relation [Gardenfors et al.88]. The consistency degree of a possibilistic CSP may be considered as an indicator of the constraints that should be suppressed for the "contraction" of a CSP upon revision. However, as an anonymous referee pointed out, that means excluding every constraint below the inconsistency degree. This is somewhat too drastic, for some of these constraints may be not "involved" in inconsistencies. This could be corrected by an adequate redefinition of the labeling compatibility, or by complete redefinition of the mesure used. However, algorithmic issues will have to be reconsidered.

- Possibility and necessity measures may be seen as specific decomposable measures [Dubois et al.82]. We think that most of this work could be easily extended to such measures (including probabilistic measures). Algorithmic issues will again have to be reconsidered.

- Possibilistic logic programming as been experimented in [Dubois et al.91a]. The integration of Possibilistic logic programming and possibilistic CSP is a first step toward Possibilistic Constraint Logic Programming.

### Acknowledgements

The author wants to thank Hélène Fargier, Jérome Lang and the anonymous referees for their fruitful comments on previous releases of this paper.